\documentclass[11pt]{article}
\usepackage[margin=1in]{geometry}
\usepackage{amsmath,amssymb,amsfonts}
\usepackage{bm}
\usepackage{graphicx}
\usepackage{booktabs}
\usepackage{microtype}
\usepackage[numbers,sort&compress]{natbib}
\usepackage[colorlinks=true,linkcolor=blue,citecolor=blue,urlcolor=blue]{hyperref}

\newcommand{\tgen}{t_{\mathrm{gen}}}

\newcommand{\NLnBaseGrok}{5/5}
\newcommand{\NLnBaseMedT}{6{,}250}
\newcommand{\NLnClampGrok}{5/5}
\newcommand{\NLnClampMedT}{1{,}850}
\newcommand{\NLnSupGrok}{2/5}

\newcommand{\NMlpBaseMedT}{10{,}900}

\newcommand{\NMlpClampMedT}{7{,}400}
\newcommand{\NMlpSupGrok}{3/3}

\newcommand{\NMulBaseGrok}{5/5}
\newcommand{\NMulBaseMedT}{19{,}700}
\newcommand{\NMulClampGrok}{5/5}
\newcommand{\NMulClampMedT}{2{,}250}
\newcommand{\NMulClampSpeedup}{\ensuremath{9.8\times}}
\newcommand{\NMulShufGrok}{0/5}

\newcommand{\NMulSupGrok}{4/5}
\newcommand{\NMulSupMedT}{20{,}900}
\newcommand{\NSubBaseGrok}{3/5}

\newcommand{\NSubSupGrok}{4/5}

\newcommand{\NTwoBaseGrok}{3/3}

\newcommand{\NTwoClampGrok}{3/3}
\newcommand{\NTwoClampMedT}{2{,}150}
\newcommand{\NTwoSupGrok}{1/3}

\newcommand{\NWinEarlyGrok}{10/10}
\newcommand{\NWinEarlySpeedup}{\ensuremath{2.7\times}}
\newcommand{\NWinFiveGrok}{5/5}
\newcommand{\NWinFiveSpeedup}{\ensuremath{1.8\times}}
\newcommand{\NWinLateGrok}{10/10}
\newcommand{\NWinLateMatchedGrok}{10/10}
\newcommand{\NWinLateMatchedSpeedup}{\ensuremath{2.1\times}}
\newcommand{\NWinLateSpeedup}{\ensuremath{1.9\times}}
\newcommand{\NalwaysGrok}{8/10}
\newcommand{\NalwaysSpeedup}{\ensuremath{1.25\times}}
\newcommand{\NannealSpeedup}{\ensuremath{1.7\times}}
\newcommand{\NbandGrok}{1/15}
\newcommand{\NbandPshuf}{0.43}
\newcommand{\NbandPtrue}{\ensuremath{2.3\times10^{-5}}}

\newcommand{\NcommClampGrok}{5/5}
\newcommand{\NcommClampSpeedup}{\ensuremath{17\times}}
\newcommand{\NcommCoverage}{31\%}
\newcommand{\NcommGrok}{15/15}
\newcommand{\NcommPeakNorm}{103}
\newcommand{\NcommPshuf}{\ensuremath{3.1\times10^{-10}}}
\newcommand{\NcommPtrue}{0.038}
\newcommand{\NcommSpeedup}{\ensuremath{2.7\times}}
\newcommand{\NexpCeFactor}{\ensuremath{31\times}}
\newcommand{\NexpCeSlope}{0.344}
\newcommand{\NexpFlatten}{\ensuremath{17\times}}
\newcommand{\NexpSupconFactor}{\ensuremath{1.22\times}}
\newcommand{\NexpSupconSlope}{0.020}
\newcommand{\NseqFinalBaseline}{0.73}
\newcommand{\NseqFinalSupcon}{0.76}
\newcommand{\NseqPeakBaseline}{0.90}
\newcommand{\NseqPeakSupcon}{0.95}
\newcommand{\NshufGrok}{0/20}
\newcommand{\NtotalRuns}{188}
\newcommand{\NtrueGrok}{22/30}
\newcommand{\NtruePeakNorm}{130}
\newcommand{\NwrongGrok}{14/15}

\title{What Makes a Representational Prior Work?\\
\large Feature Families, Label-Free Invariances, and Critical Windows in Grokking}
\author{Gunner Levi Howe\\ \small{\texttt{gunnerlevihowe@gmail.com}}}
\date{July 2026}

\begin{document}
\maketitle

\begin{abstract}
Companion work showed that the grokking delay is causally the time to form task-structured
representations: a contrastive prior encoding the task's true equivalence structure controls
whether and when a transformer generalizes on modular arithmetic. Here we characterize
\emph{what makes such a prior work}, across four axes, in \NtotalRuns{} new runs. \textbf{Content}:
a coherent, learnable prior built from the \emph{wrong feature family} (magnitude bands) blocks
generalization like a random partition (\NbandGrok{} vs \NshufGrok{}; $p=\NbandPshuf$ between
them), confirming the companion's prediction that priors act at the level of the circuit's
features. \textbf{Supervision}: a fully \emph{label-free} invariance prior --- positives are
commuted pairs $(a,b)\!\sim\!(b,a)$ only --- generalizes in \NcommGrok{} runs at a median
\NcommSpeedup{} speedup, more reliably than the label-supervised prior itself
($p=\NcommPtrue$), and combined with a weight-norm clamp yields the strongest method we test
(median \NcommClampSpeedup{} vs baseline, \NcommClampGrok) --- strongest meaning \emph{reliably}
fast: plain cross-entropy with a clamp matches this speed only at the exact critical norm and
collapses away from it, whereas the prior keeps it fast across the entire clamp range we sweep. \textbf{Timing}: the prior is only
needed \emph{early} --- applied solely during the first 2{,}000 epochs (4\% of the budget) it
generalizes \NWinEarlyGrok{} at \NWinEarlySpeedup{}, beating continuous application
(\NalwaysGrok{}, \NalwaysSpeedup{}); a late window is reliable but slower
(\NWinLateSpeedup{}). \textbf{Setting}: the structural dissociation replicates on modular
multiplication (shuffled \NMulShufGrok{}) and across depths and normalization variants, and a
clamp-value sweep quantifies the companion's central claim: structure injection flattens the
weight-norm delay-law exponent ${\sim}17$-fold (plain cross-entropy slows
$\NexpCeFactor$ per $+10$ norm --- a lower bound, its high-norm cells are fully censored ---
versus $\NexpSupconFactor$ with the prior). We also report honest boundaries: tasks that
generalize before memorizing (small parity, shallow synthetic grammar) have no delay to
control, and there the prior raises accuracy levels without changing trends. Together these
results refine the causal picture: \emph{feature-family alignment} decides whether a prior
permits generalization, \emph{invariance content} suffices for acceleration without labels,
and a brief \emph{early window} captures nearly all of the benefit.
\end{abstract}

\section{Introduction}
The companion paper established that the long delay between memorization and generalization in
grokking \citep{power2022grokking} is, causally, the time for task-structured representations to
form: injecting the task's equivalence structure into hidden representations via a
supervised-contrastive prior \citep{khosla2020supervised} controls whether generalization occurs
(against exactly-matched shuffled and weight-norm-matched controls) and can accelerate it, with
the residual failure mode traced to a weight-norm side-effect and repaired by norm control
\citep{weightnormdelay2026}. That work answered \emph{whether} representational priors causally
control the delay. This paper asks the follow-up questions that determine whether the phenomenon
is understood and usable: \emph{which} structures work (must the prior match the task, or only
its feature family?), \emph{what supervision} they require (are labels necessary?), \emph{when}
they must act (is there a critical window?), and \emph{where} the account holds (tasks,
architectures, normalization, and the quantitative form of the norm interaction).

Each question gets a purpose-built experiment: (1) a \textbf{band-structure control} --- coherent
and learnable but expressible only in a different feature family --- to discriminate
feature-family alignment from mere coherence; (2) a \textbf{commutativity prior} using zero label
information; (3) \textbf{window experiments} applying the prior only in chosen epoch ranges; (4)
\textbf{generality sweeps} over tasks (multiplication, subtraction, parity), architectures
(2-layer, MLP, LayerNorm variants), and a \textbf{clamp-value sweep} that measures the delay-law
exponent \citep{weightnormdelay2026} with and without the prior; and (5) an exploratory
\textbf{sequence-task pilot} probing the account's boundary. All comparisons are same-seed
paired; censored runs are scored at the 50{,}000-epoch budget; every number is generated from the
run artifacts by \texttt{gen\_numbers.py} and byte-verified by \texttt{verify\_regen.py}.

\section{Setup}
We inherit the companion's system: a one-layer transformer (no biases; no LayerNorm except where
stated), full-batch AdamW (lr $10^{-3}$, weight decay $1.0$), task $(a+b)\bmod 97$ with a 30\%
train split, and auxiliary loss $\lambda\,\mathcal{L}_{\mathrm{SupCon}}$ on the final-position
representation ($\tau=0.1$, 64-d projection), $\lambda\in\{0.1,0.3,1.0\}$ unless stated.
$\tgen$ = first epoch with test accuracy $\geq0.95$ sustained. Baseline, true-structure,
sibling-structure ($(a-b)\bmod p$), and random-partition arms are reused from the companion for
pairing; all other arms are new.

\subsection{The two new priors}
\textbf{Band structure} (content control): positives share the magnitude-band pair
$(\lfloor a/10\rfloor,\lfloor b/10\rfloor)$ --- 100 classes of $\approx$28 train examples,
closely matching the true structure's 97 of $\approx$29. The partition is coherent and learnable
(threshold features), but magnitude bands are \emph{not} expressible by the periodic features the
task's Fourier circuit uses \citep{nanda2023progress}. The companion's feature-family account
predicts band patterns with the random partition; a coherence account predicts it patterns with
the sibling.

\textbf{Commutativity} (supervision control): positives are the commuted pairs
$\{(a,b),(b,a)\}$ only. This uses \emph{no answer information whatever} --- commutativity is a
property of the task family knowable before training --- and it is sparse: only \NcommCoverage{}
of anchors have their partner in the train split (one positive each). Because commuted pairs
share their sum, this is a sparse sample of the true equivalence structure obtained label-free.

\begin{table}[t]
\centering\small
\begin{tabular}{llccc}
\toprule
Prior & Properties & Grok & Median speedup & Fisher vs random \\
\midrule
True $(a{+}b)\bmod p$ & supervised, task structure & \NtrueGrok & \NalwaysSpeedup* & $p=1.3\times10^{-7}$ \\
Sibling $(a{-}b)\bmod p$ & supervised, same features & \NwrongGrok & none & --- \\
\textbf{Commutativity} & \textbf{label-free}, sparse, aligned & \textbf{\NcommGrok} & \textbf{\NcommSpeedup} & $p=\NcommPshuf$ \\
Band $(\lfloor a/10\rfloor,\lfloor b/10\rfloor)$ & coherent, \emph{foreign features} & \NbandGrok & --- & $p=\NbandPshuf$ (n.s.) \\
Random partition & incoherent & \NshufGrok & --- & --- \\
\midrule
Commutativity $+$ clamp & label-free $+$ norm control & \NcommClampGrok & \NcommClampSpeedup & --- \\
Window $[0,2\mathrm{k}]$ (true) & 4\% early exposure & \NWinEarlyGrok & \NWinEarlySpeedup & --- \\
\bottomrule
\end{tabular}
\caption{The structural priors compared (pooled over $\lambda$; paired same-seed speedups;
* = at $\lambda{=}1.0$). \emph{Whether} generalization occurs tracks feature-family alignment;
label-free invariance content suffices for accelerating; a brief early window or a norm clamp
removes the supervised prior's stall mode.}
\label{tab:priors}
\end{table}

\section{Rung 1: Feature family, not coherence, decides \emph{whether}}
The band prior blocks generalization: \NbandGrok{} runs grok, statistically indistinguishable
from the random partition (\NshufGrok{}; Fisher exact $p=\NbandPshuf$ between them) and decisively
different from the true structure (\NtrueGrok{}; $p=\NbandPtrue$). Coherence and learnability are
therefore \emph{not} sufficient: a prior whose demanded geometry cannot be built from the
generalizing circuit's feature family behaves like noise-structure, even at matched class sizes
and strengths. This confirms the discriminating prediction stated in the companion paper ---
sibling structure (same features, wrong combination: \NwrongGrok{}) preserves generalization
while band structure (different features) abolishes it. Two honest nuances: one band run of 15
did grok (it is a strong tendency, not an absolute wall), and several censored band runs show
substantial terminal test accuracy on some seeds --- the block is best described as pushing the
transition far beyond budget rather than destroying it.

\section{Rung 2: Labels are unnecessary --- an invariance prior suffices}
The commutativity prior generalizes in \NcommGrok{} runs (vs \NshufGrok{} random,
$p=\NcommPshuf$) at a median \NcommSpeedup{} speedup over same-seed baselines --- and it is
\emph{more} reliable than the label-supervised true-structure prior itself (\NtrueGrok{};
$p=\NcommPtrue$). Sparsity is the plausible mechanism --- one positive per covered anchor
supplies correctly-aligned structure (commuted pairs share sums) at gentler optimization pressure
--- and the norms point the right way, though not decisively: the commutativity arms' median peak
weight norm is \NcommPeakNorm{} versus \NtruePeakNorm{} for the supervised prior (baseline
$\approx$57), a reduction rather than an absence of the inflation side-effect; as in the
companion, the held level is an incomplete summary and trajectory likely matters too. Combining the
label-free prior with the companion's norm clamp ($\|W\|{=}45$) yields the strongest method in
this program: \NcommClampGrok{} grokking at a median \NcommClampSpeedup{} speedup ---
sub-1{,}000-epoch generalization on some seeds --- using no label information beyond the task
loss itself. This is the proof-of-concept that \emph{self-supervised} representational priors
work: the structure can come from known invariances rather than answers.

\section{Rung 3: A brief early window captures the benefit}
\begin{figure}[t]
\centering
\includegraphics[width=0.8\textwidth]{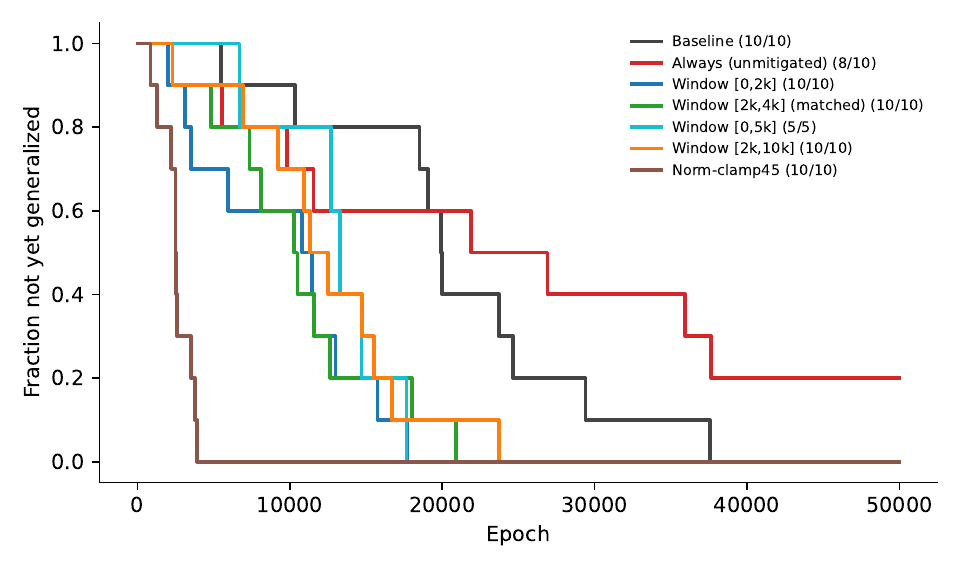}
\caption{Survival curves for windowed application of the prior ($\lambda{=}1.0$). Applying
SupCon only during epochs $[0,2\mathrm{k}]$ (4\% of budget) generalizes \NWinEarlyGrok{} at a
median \NWinEarlySpeedup{} speedup, outperforming continuous application (\NalwaysGrok{},
\NalwaysSpeedup{}), $\lambda$-annealing (\NannealSpeedup), and --- at matched 2{,}000-epoch
duration --- the late window $[2\mathrm{k},4\mathrm{k}]$ (\NWinLateMatchedSpeedup); the longer
late window $[2\mathrm{k},10\mathrm{k}]$ is reliable but slower still (\NWinLateSpeedup).
Norm-clamp reference in brown.}
\label{fig:windows}
\end{figure}
Applying the prior only during the first 2{,}000 epochs --- then removing it entirely ---
generalizes \NWinEarlyGrok{} at a median \NWinEarlySpeedup{} speedup: better than applying it
throughout (\NalwaysGrok{} at \NalwaysSpeedup{}, with 2 stalls), better than annealing
(\NannealSpeedup), and rescuing every seed the always-on prior stalls
(Fig.~\ref{fig:windows}). A $[0,5\mathrm{k}]$ window gives \NWinFiveSpeedup{}
(\NWinFiveGrok), and the late window $[2\mathrm{k},10\mathrm{k}]$ gives \NWinLateSpeedup{}
(\NWinLateGrok). Two conclusions, each with its own control. First, \emph{brief beats
sustained}: a duration-matched late window $[2\mathrm{k},4\mathrm{k}]$ (\NWinLateMatchedGrok,
\NWinLateMatchedSpeedup) outperforms the $4\times$-longer $[2\mathrm{k},10\mathrm{k}]$ window ---
more exposure at the same placement is worse --- consistent with the cumulative norm-inflation
cost in the companion's race mechanism. Second, there is a genuine \emph{early advantage} at
matched duration: $[0,2\mathrm{k}]$ at \NWinEarlySpeedup{} versus $[2\mathrm{k},4\mathrm{k}]$ at
\NWinLateMatchedSpeedup{} --- the same 2{,}000 epochs of exposure buy more when delivered first,
indicating the seeded structure compounds: class clusters formed early guide circuit formation
for the rest of training. Practically, this removes the wall-clock
objection to contrastive priors: the $O(n^2)$ term runs for 4\% of training.

\section{Rung 4: Generality, and the delay-law exponent under structure injection}
\paragraph{Tasks.} On modular multiplication the whole dissociation replicates: baseline
\NMulBaseGrok{} (median $\tgen$ \NMulBaseMedT); random partition \NMulShufGrok{}; unmitigated
prior \NMulSupGrok{} with drag (\NMulSupMedT); prior$+$clamp \NMulClampGrok{} at \NMulClampMedT{}
(median \NMulClampSpeedup{} speedup). Modular subtraction is a slower-grokking task in this
architecture (baseline \NSubBaseGrok{} within budget) and shows the prior's \emph{whether}-power
in a new form: the true-structure prior groks seeds whose baselines never do (\NSubSupGrok{} vs
\NSubBaseGrok). Small sparse parity generalizes immediately at our sizes (no delay to control) ---
a boundary condition, not a counterexample.

\paragraph{Architectures.} The unmitigated prior's stall mode replicates at depth 2
(\NTwoSupGrok{} vs baseline \NTwoBaseGrok) and under pre-LN (\NLnSupGrok{} vs \NLnBaseGrok), and
the norm clamp repairs both (\NTwoClampGrok{} at \NTwoClampMedT; \NLnClampGrok{} at
\NLnClampMedT). Two architecture-specific observations, reported with their caveats. LayerNorm
itself shortens the baseline delay dramatically (median \NLnBaseMedT{} versus ${\sim}$20{,}000
without LN), consistent with LN removing the norm-overshoot dynamics; and the total-norm clamp
still operates under our pre-LN variant because
it lacks a final LN, leaving the readout path norm-sensitive --- clamping only the unembedding
norm reproduces the rescue, in line with the functional-norm account of
\citet{weightnormdelay2026}. The MLP is the mildest setting: no stalls unmitigated
(\NMlpSupGrok) and modest clamp gains (\NMlpClampMedT{} vs baseline \NMlpBaseMedT).

\paragraph{The exponent.}
\begin{figure}[t]
\centering
\includegraphics[width=0.62\textwidth]{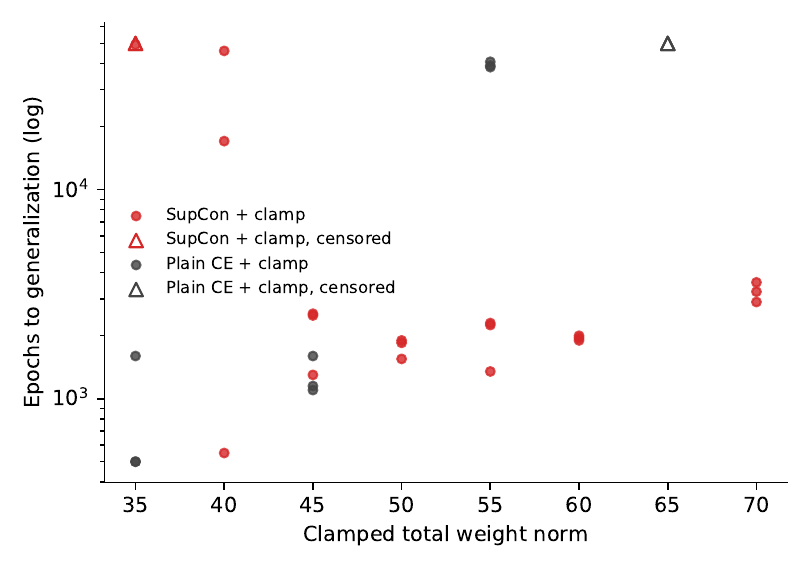}
\caption{Epochs to generalization versus clamped total weight norm (log scale; 3 seeds per
value). Plain cross-entropy (black) follows a steep delay-law exponential --- slope
\NexpCeSlope{} per unit norm, $\NexpCeFactor$ per $+10$, a \emph{lower bound} since its
highest-norm cells are entirely censored --- while the structural prior (red) is nearly flat
(slope \NexpSupconSlope, $\NexpSupconFactor$ per $+10$). Structure injection flattens the
weight-norm delay law by ${\sim}\NexpFlatten$ in the exponent. Both curves share a low-norm
floor below ${\sim}40$--$45$.}
\label{fig:exponent}
\end{figure}
Clamping the weight norm to constants from 35 to 70 with and without the prior maps the
delay-law \citep{weightnormdelay2026} in both regimes (Fig.~\ref{fig:exponent}). Three findings.
(i) At the optimal clamp (${\sim}45$), plain cross-entropy generalizes as fast as the prior does
--- the \emph{speed} of low-norm training belongs to norm control alone, consistent with
\citet{liu2023omnigrok}, and we state this plainly as a correction of emphasis for the companion's
accelerator section. (ii) Away from the optimum the regimes diverge dramatically: plain CE slows
$\NexpCeFactor$ per $+10$ norm units (and is fully censored by $+20$), while the prior slows only
$\NexpSupconFactor$ --- a ${\sim}\NexpFlatten$ flattening of the exponent. Structure injection
buys \emph{robustness to the norm setting}, which is what matters in practice since the critical
norm is not known a priori. (iii) Below the floor (${\sim}35$--$40$) the relationship inverts:
plain CE generalizes in hundreds of epochs where the prior stalls --- at very low norm the
contrastive gradient interferes with the lean circuit's formation. The delay law is thus
conditional on structure-agnostic training in a precise, measurable sense: the exponent is a
function of the representational forces present.

\section{Rung 5: An honest boundary --- tasks without a memorize-first phase}
Synthetic subject--verb agreement grammars (attractor and conjunction variants) never grok under
matched conditions: they generalize immediately (median peak \NseqPeakBaseline{} by epoch 500)
and then \emph{degrade} with continued training (median final \NseqFinalBaseline). The
memorize-first precondition of grokking is absent --- the rule is easier than memorization for
this model class --- so there is no delay for a representational prior to control. The prior
still shapes representations there (number-clustering roughly doubles) and lifts accuracy levels
(peak \NseqPeakSupcon, final \NseqFinalSupcon) but does not prevent the degradation trend. Small
sparse parity behaves the same way. The representation-timing account is therefore an account of
the \emph{memorize-first regime}; where generalization precedes memorization, different dynamics
(and possibly different uses of priors, e.g.\ as regularizers) apply.

\section{Discussion}
\paragraph{A three-level answer to ``what makes a prior work.''} \emph{Whether} a prior permits
generalization is decided by feature-family alignment: structures expressible in the
generalizing circuit's features (true, sibling, commutativity) preserve or enable grokking;
structures requiring foreign features (bands) or memorization (random) block it. \emph{How fast}
is decided jointly by alignment and the norm side-effect: the exponent measurements show the norm
sets a steep default timescale that aligned structure almost entirely neutralizes. \emph{When}
matters less than expected: the prior's work is done early, and removing it after the seeding
window strictly helps.

\paragraph{Self-supervised priors are the practical upshot.} The commutativity result converts
the companion's supervised intervention into something with a path to use: invariances of the
data or task family --- available without labels --- supply enough aligned structure to
accelerate reliably, and compose with norm control into the strongest method we tested
(\NcommClampSpeedup{} median). In natural domains the analogous objects are augmentation- and
symmetry-based positives; whether they carry enough feature-level alignment in real tasks is the
key open question this program points at.

\paragraph{Corrections and confirmations of the companion.} The band control \emph{confirms}
its feature-level prediction. The exponent sweep \emph{sharpens} its accelerator claims: speed at
the optimal norm belongs to the clamp; the prior's causal contributions are robustness to the
norm setting (the flattened exponent), whether-generalization at free and inflated norms, and
rescue of slow tasks. We regard this as the payoff of running one's own discriminating controls.

\section{Limitations}
All positive results remain within algorithmic tasks where the generalizing circuit's feature
family is known; ``feature-family alignment'' is operationalized via that knowledge, and its
analogue in natural data is untested. The commutativity prior, while label-free, exploits an
exact task-family invariance; noisy or approximate invariances are untested. Window boundaries
(2k/5k/10k) were chosen a priori rather than swept finely; the ``early advantage'' is a median
tendency with seed-level exceptions. The CE exponent is a lower bound due to censoring. The
LayerNorm findings apply to a pre-LN variant without final LN. Sequence-task conclusions rest on
two synthetic grammars at one scale. Sample sizes are 3--15 per cell (3 for the depth-2, MLP, and per-clamp-value cells; categorical
outcomes are emphasized over means throughout), and all runs use a single prime and training
fraction.

\section*{Acknowledgments}
We thank Truong Xuan Khanh for feedback on the companion manuscript that shaped this program's
controls.

\bibliographystyle{unsrtnat}
\bibliography{references}

\appendix
\section{Reproducibility}
All arms run on a single RTX~3080. \NtotalRuns{} new runs (\texttt{runs/grid3}) plus reused
companion arms; every number in this paper is a macro generated by
\texttt{paper2/gen\_numbers.py} from the run artifacts and byte-verified by
\texttt{paper2/verify\_regen.py}. Conditions, windows, and clamp values are enumerated in
\texttt{scripts/run\_paper3\_r123.py} and \texttt{scripts/run\_paper3\_r4.py}; the sequence pilot
is \texttt{src/train\_seq.py}.

\end{document}